\pdfoutput=1

\documentclass[11pt,a4paper]{article}
\usepackage[hyperref]{acl2019}
\usepackage{times}
\usepackage{latexsym}
\usepackage{algorithm,algorithmic}
\usepackage{graphicx}
\usepackage{bm}
\usepackage{url}
\aclfinalcopy

\title{Blind signal decomposition of various word embeddings based on join and individual variance explained (JIVE)}

\author{Yikai Wang \\
  Department of Biostatistics\&Bioinformatics\\ Emory University \\
  \texttt{ywan566@emory.edu} \\\And
  Weijian Li \\
  Department of Computer Science\\ Emory University\\
  \texttt{weijian.li@emory.edu} \\}

\date{}

\begin{document}
\maketitle
\begin{abstract}
  In recent years, natural language processing (NLP) has become one of the most important areas with various applications in human's life. As the most fundamental task, the field of word embedding still requires more attention and research. Currently, existing works about word embedding are focusing on proposing novel embedding algorithms and dimension reduction techniques on well-trained word embeddings. In this paper, we propose to use a novel joint signal separation method - JIVE to jointly decompose various trained word embeddings into joint and individual components. Through this decomposition framework, we can easily investigate the similarity and difference among different word embeddings. We conducted extensive empirical study on word2vec, FastText and GLoVE trained on different corpus and with different dimensions. We compared the performance of different decomposed components based on sentiment analysis on Twitter and Stanford sentiment treebank. We found that by mapping different word embeddings into the joint component, sentiment performance can be greatly improved for the original word embeddings with lower performance. Moreover, we found that by concatenating different components together, the same model can achieve better performance. These findings provide great insights into the word embeddings and our work offer a new of generating word embeddings by fusing.

\end{abstract}

\section{Introduction}

Nowadays, with the advance of technology and information system, natural language processing (NLP) has become one of the most important areas in both industry and scientific research \citep{turney2010frequency,cambria2014jumping}. Applications of NLP have been involved in many different areas, such as language translation, keyword searching, machine comprehension, sentiment analysis, user language interface, cross-language information retrieval, speech recognition and other language related text mining tasks \citep{kim2014convolutional,salton1988term,yin2017deepprobe,xu2015show,mikolov2013distributed,lai2015recurrent,dos2014deep}. The core task of NLP is to make computer system understand and manipulate natural human's language to perform various tasks. However, this is a very challenging task. 

First of all, the fundamental question of any kind of NLP tasks is how to deal with a large number of words (i.e. the smallest element in one sentence or passage) in the vocabulary list \citep{mikolov2013distributed}. For example, Oxford English Dictionary, published in 1989, contains full entries for 171,476 words in current use \citep{murray1933oxford}. Initially, research built language processing system based on N-gram language model or bag-of-word concept using word counts or frequency, i.e. tf-idf \citep{aizawa2003information,cambria2014jumping}. Obviously, this type of approach cannot fully capture the rich information of human natural language and has limited performance. So far, one the most fundamental and successful achievement in NLP is the concept of word embedding, which aims at learning a vector space representation of each single word in low dimensional space. Some famous and well-performed word embedding algorithms includes Word2Vec \citep{mikolov2013distributed}, FastText \citep{bojanowski2017enriching} and GLoVE \citep{pennington2014glove}. Each of these methods try to learn a relatively low-dimensional vector representation of each word based on the co-occurace pattern of words in the corpus in an unsupervised manner. For example, Word2Vec trains the embedding of the word by using its surrounding window of context words as predictor or outcomes. Therefore, the nearby words tend to have similar word embeddings. Moreover, as for FastText, it further makes use of the information in the character level (sub-element consists of word), and GLoVE learns the embedding based on decomposing the global co-occurrence matrix of words. Nowadays, these embedding algorithms are serving as the foundation for NLP, and built upon these algorithm, a growing number of machine learning and deep learning models are being developed in recent years and achieve great success in many areas. 

However, although word embedding algorithm facilitates the model development of NLP, limited work or research has been conducted to analyze or understand the output of these embedding algorithms. Most work simply treats them as 'hyper-parameter' and changes the embedding algorithm based on their performance on available datasets. However, given the importance of the word embedding in NLP, we argue that more efforts are needed for investigating the difference and the meaning of the output of word embedding algorithm. Unfortunately, given the dimensionality, evaluating the difference or similarity of various word embeddings is very challenging. Existing works are only focusing on conducting dimension reduction on one word embedding output instead of understanding the common and individual part of various word embeddings \citep{levy2014neural,iacobacci2015sensembed,yin2018dimensionality}. For example, \citet{yin2018dimensionality} provide a theoretical understanding of word embedding to guide the selection the dimension. \citet{iacobacci2015sensembed} aims to further transfer the estimated embeddings to capture more information. Different from these existing works, we consider multiple word embeddings and utilize joint signal decomposition approach to map them into joint and individual components for following analysis. 

In this paper, we propose to jointly study the word embeddings learned from different algorithms by decomposing them into common and individual components based on the joint signal separation method - JIVE \citep{lock2013joint,o2016r,feng2018angle}. Specifically, JIVE can decompose different word embeddings into the summation of shared column space and the orthogonal embedding-specific space. Based on this outcome, we can answer several fundamental questions: 1) what is the similarity among different word embedding algorithms; 2) what is the difference between two word embedding algorithms. In summary, we study 3 important questions for word embedding based on JIVE analysis on word embeddings: 1) whether the information captured by word embedding increases linearly with the dimension of embedding space; 2) what is the difference and similarity for word embeddings trained with the same algorithm but different corpus; 3) what is the difference and whether there is similarity between different word embedding algorithms. Moreover, there are other relevant approaches in the literature for joint decomposition including canonical correlation analysis (CCA) \citep{hotelling1992relations}, partial least squares (PLS) \citep{helland2014partial}, multi-level functional PCA \citep{di2009multilevel}, parallel and joint independent component analysis \citep{eichele2008unmixing,moosmann2008joint,wang2019hierarchical,lukemire2018hint, mejia2020template,wang2020locus}. These methods cannot provide joint and individual information like JIVE and hence is not what we need. Furthermore, our work can provide great insights for better understanding the word embeddings and can potentially give directions for generating better word embeddings or guidances on how to compress/control the information in word embeddings. 

This paper is organized as follows: section \ref{method} introduces the word embedding algorithm being considered and the model specification of JIVE. Section \ref{design} discusses the design of our experiments and section \ref{result} presents the results as well as the discussion. Section \ref{conclu} discusses the finding in this paper and make conclusion, and section \ref{future} talks about limitation and the potential future direction.

\section{Methodology}\label{method}

In this section, we will first introduce three commonly used word embedding algorithms and then formally define the model specification, hyper-parameter selection and algorithm for JIVE.

\subsection{Word Embedding}
We will briefly introduce three commonly used word embedding algorithms being considered in this paper: Word2Vec \citep{mikolov2013distributed}, fastText \citep{bojanowski2017enriching} and GloVe \citep{pennington2014glove}. 

\textbf{Word2Vec} is one of the first word embeddings that is applied in NLP tasks. It is a predictive model trained on the context of a word in a large corpus. During the training, either CBOW or skip-gram can be used to model the objective function. CBOW uses the context to predict he center word, whereas skip-gram uses the center word to predict he context. The objective function aims to minimize the loss of this prediction given the word vectors. Either negative sampling or hierarchical softmax can be used to reduce the computational complexity. 

\textbf{FastText} is trained in a very similar way as Word2Vec, except that fastText is trained with character level n-grams instead of word level. With this formation, the model can better capture sub-word information like prefix and suffix, which also contains semantic meanings. Because of this nature, fastText is good at handling unseen words and misspelling. 

\textbf{GloVe} is a count-based model. It first construct a large co-occurrence matrix based on the frequency of each word in its context. Matrix factorization then is used to reduce the dimensionality of this large matrix to gain a vector representation for each word. The objective function aims to minimize the reconstruction loss based on this dimension-reduced matrix. 

As we discussed, these word embedding methods all achieved great success in different applications but little is known for the difference and similarity among them. In the next section, we will introduce a novel statistical signal decomposition technique which can jointly decompose different word embeddings into joint and individual components for further analysis. 

\subsection{Joint and Individual Variance Explained}\label{jive}
Joint and Individual Variance Explained (JIVE) is a statistical approach for decomposing multiple data modalities into joint component and modality-specific components. JIVE is originally proposed by \citet{lock2013joint} for decomposing different features on the same subject (gene and RNA). In this section, we will provide the formal model specification of JIVE and its assumption. 

First, denote $X_i \in \mathcal{R}^{p_i \times n}$ to be the $i$th dataset from same $n$ subjects with different feature spaces, $i=1,...,K$. For example, in \citet{lock2013joint}, $X_1$ denotes the gene expression data same and $X_2$ denotes miRNA data. In our case, $X_i$'s will denote different word embeddings with $n$ to be the number of distinct words and $p_i$ denotes the dimension for word embedding. The unified model of JIVE will be:
\begin{equation}
    X_i = J_i + A_i + E_i, 
\end{equation}

\begin{figure}
    \centering
    \includegraphics[scale=0.5]{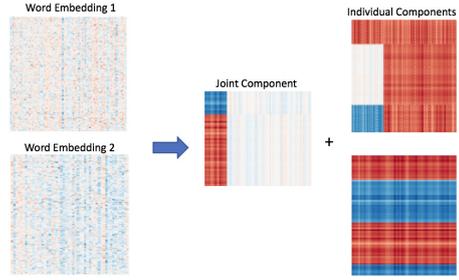}
    \caption{Illustration of JIVE analysis for jointly decomposing word embeddings}
    \label{fig:jive}
\end{figure}
where $J_i$ captures the joint information among all data modalities and $A_i$ denotes the modality-specific information, and $E_i$ denotes residuals with expectation of zero. To ensure the identifiability of JIVE, we have the following constraints: $J_i A_i' = 0$ for $i=1,...,K$. The overal summary of JIVE is summarized in Figure \ref{fig:jive}. 

Specifically, all joint components $J_i$ shares the same row space and can be represented as $J_i = B_i J$ with $J \in \mathcal{R}^{r \times n}$ to be the basis of the same row space and $B_i$ denotes the linear transformation and is assumed to satisfy that $B_i'B_i = I_{p_i}$. For individiual component, we can also re-write it as $A_i = D_i H_i$ with $D_i'D_i = I_{p_i}$ and $JH_i'=0$. Therefore, JIVE is a natural extension of PCA for multiple modalities where $J$ and $H_i$ can be viewed as the principle components for $X_i$. In summary, we have 
\begin{equation}
    X_i = B_i J + D_i H_i + E_i.
\end{equation}

\subsection{JIVE on Word Embeddings}
Section \ref{jive} introduce the definion of JIVE. In this section, we will present more details about JIVE analysis on word embeddings. 

\subsubsection{Joint Rank Selection}

First, the hyper-parameter in JIVE is the rank of joint component $J_i$ or the dimension of $J$, which is $r$. It controls the amount of information shared acorss different word embeddings and its value can significantly affect the conclusion. In the literature, there are three commonly used approaches for selection of $r$: permutation test \citep{lock2013joint}, BIC based selection \citep{o2016r, wang2020locus} and a theoretical selection approach based on Wedin bound \citep{feng2018angle}. Although permutation based approach tends to be the most robust method, its application for word embedding analysis is not desired because the computation burden is very expensive because of the sample size, number of word in dictionary. Moreover, BIC method requires the specification of probability distribution on word embeddings. However, different embedding method has different distributional assumption and hence the BIC based method can be very complicated.  Therefore, we select to use Wedin bound based selection approach for JIVE analysis of word embeddings. 

\subsubsection{Algorithm for JIVE}

Second, in this section, we present the algorithm for estimating the unknown parameters in JIVE. Given the limited resources of JIVE in python comunity, we followed the algorithm proposed in \citet{lock2013joint} and implemented JIVE in Python 3.0. We proposed an efficient approximate estimation method used for initial starting point. The method for 2 word embeddings for estimating the initial value and the iterative algorithm for JIVE are summarized in Algorithm 1. It can be easily extended to the case with more than 2 word embeddings.

\begin{algorithm}\label{algo_jive}
\caption{ JIVE with two word embeddings}
\begin{algorithmic}
    \STATE {\bfseries Inputs}: Concatenated embedding matrix $\bm X = [\bm X_1', \bm X_2']'$ of dimension $p_1+p_2 \times N$, rank of joint component $r$, rank of individual terms $r_1, r_2$. \\
    \STATE {\bfseries Initial values}:  \\
    \STATE\quad 1. Conduct SVD on $X = U$ diag $(\Lambda_1,...,\Lambda_{p_1+p_2}) V'$, set $J^{(0)} = U$ diag $(\Lambda_1,...,\Lambda_r,0,..,0) V'$. \\
    \STATE\quad 2. Conduct 2 separate SVD on first $p_1$ and last $p_2$ rows of $X-J^{(0)}$, and only takes first $r_i$ terms for $A_i$, $A^{(0)} = [A_1',A_2']'$. 
    \STATE\quad 3. Calculate $R^{(0)} = \|X-J^{(0)}-A^{(0)}\|_2^2$.
 \STATE {\bfseries Estimation} \\
   \REPEAT
       \STATE 1. Condition on $A^{(t)}$, update $J$ of rank $r$ to minimize $\|X-J-A^{(t)}\|_2^2$. 
       \STATE 2. Condition on $J^{(t+1)}$, update $A_1$ of rank $r_1$ and $A_2$ of rank $r_2$ to minimize $\|X-J^{(t+1)}- A\|_2^2$
       \STATE 3. Calculate $R^{(t+1)} = \|X-J^{(t+1)}-A^{(t+1)}\|_2^2$
\UNTIL{\textbf{convergence}, i.e. $\frac{R^{(t+1)}-R^{(t)}}{R^{(t)}} <\epsilon$}
\end{algorithmic}
\end{algorithm}

The word embeddings $X_i$ in Algorithm 1 is assumed to be de-meaned at each row/feature and re-scaled to have a standard deviation of 1 for the whole embedding, i.e. $X_i = X_i / \|X_i\|$ where where $\|\cdot\|$ defines the Frobenius norm. 

The initial estimation follows a 2-step procedure which first conduct a single PCA on concatenated data matrix to extract the joint component. In the second step, it removes the estimated joint component from the data matrix and conductes 2 separated PCA for each modalities. The esimation algorithm for JIVE is adopted from \citet{lock2013joint}. We will check the algorithm convergence in section \ref{result}.

\section{Experimental Design} \label{design}

In this section, we will introduce the details about our experimental design. We first introduce the datasets used in this paper and then the 3 different downstream tasks being considered (summarized in Table 1). Finally, we will discuss about the motivation and assumptions for each experiment settings.

\begin{figure}
    \centering
    \includegraphics[scale=0.5]{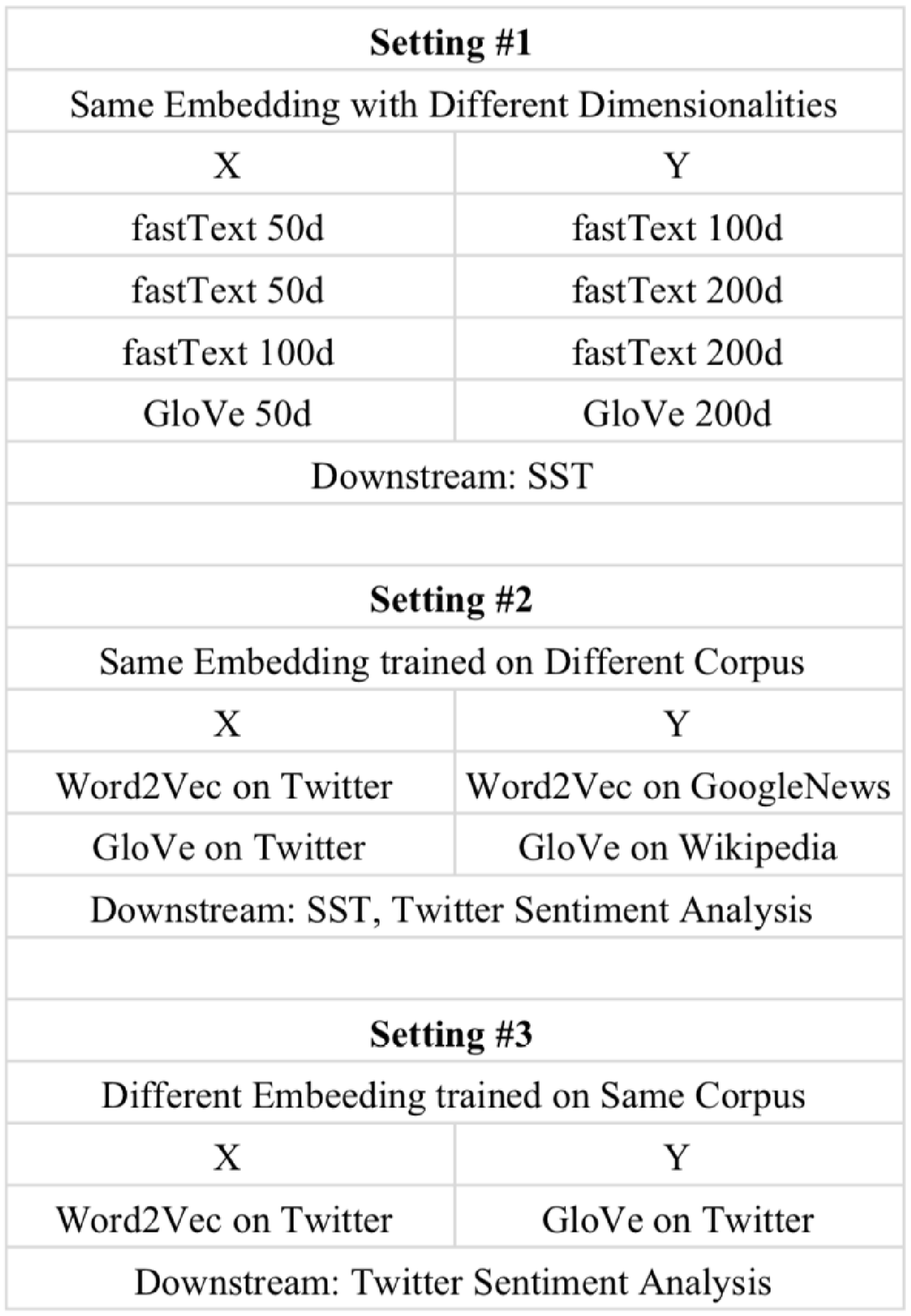}
    \caption*{Table 1. A brief summary of experimental design}
    \label{tab:design}
\end{figure}

\subsection{Downstream Task and Datasets}
For the downstream task, we wanted to choose a task of which the results and performance depends largely on the quality of the word embeddings itself, rather than on the structure of the neural network model. Our reasoning behind this was that if the neural network itself is too complicated, we will not be able to tell the performance differences between the word embeddings, because a complicated model might compensate the negative effects caused by bad embeddings. To this end, we chose sentiment analysis as our downstream task, because the prediction can be affected by the meaning of a few keywords, in which case the quality of word embeddings is decisive. 

We chose two datasets for the sentiment analysis task. One is Stanford Sentiment Treebank \citep{socher2013recursive}, the other is \hyperlink{https://www.kaggle.com/c/twitter-sentiment-analysis2/overview}{Twitter sentiment analysis dataset}. SST dataset contains movie reviews crawled from Rotten Tomato. The language is comparatively more formal than Twitter. The training set for SST has 8539 sentences, and testing set has 1101 sentences. The vocabulary size is 18193. The target has 5 categories from 0 to 4, ranging from most negative to most positive. The Twitter dataset contains 50248 tweets crawled from Twitter, of which half were used for training and half for testing. The vocabulary size for Twitter is 24604. Different from SST, the Twitter dataset is a binary classification problem with only 2 sentiments, positive and negative. 

\subsection{Same Embedding with Different Dimensionalities}\label{sEdD}
We explored the effects brought by the difference in dimensionalities of the same embedding model. Intuitively, the model with higher dimensionality should encode more information than the model with lower dimensionality. Besides, the model with higher dimensionality should encode all the information that is already encoded by the model with lower dimensionality. With the joint and individual matrices generated by JIVE, we were able to test our intuition. 

For this experiment, we first ran JIVE on 2 word embeddings X and Y which have different dimensionalities. The generated joint matrix represented the common information encoded by the two embeddings, and had the shape of vocabulary size by joint rank. The 2 individual matrices represented the distinctive information that is not shared between the two embeddings. And this part of information should be the potential cause of the different performances showed by the two embeddings. With the joint and individual matrices, we were able to directly use them as word embeddings to conduct downstream tasks. We also concatenated 1) Joint matrix with individual matrix of X; 2) Joint matrix with individual matrix of Y; 3) Individual matrix of X and that of Y and 4) Joint matrix and the 2 individual matrices. This gave us 4 more word embeddings. The performance of each word embedding gave us insights on what information are encoded by the original embeddings. 

Specifically, we conducted experiments on the following groups of embeddings on the SST dataset: fastText 50d with fastText 100d, fastText 50d with fastText 200d, fastText 100d with fastText 200d and GloVe 50d with GloVe 200d. 

\subsection{Same Embedding Trained on Different Corpus}\label{sEdC}
We also explored the effects brought by the different training corpus of the same embedding algorithms. The intuition was that the word embedding trained on a formal language corpus, e.g. Wikipedia, might result in better performance on formal language downstream tasks. On the other hand, the word embedding trained on a colloquial corpus,e.g. Twitter, could perform better on colloquial language downstream tasks. Even though the word embedding model was generated by the same algorithm, the information they encoded could differ drastically. Meanwhile, we wanted to explore the JIVE's effects on these embeddings. Similar to the precious setting, we will use the joint and the 2 individual matrices to generate 4 more new embeddings. Finally, we will use this embeddings to conduct 2 downstream tasks: SST and Twitter sentiment analysis. The reason behind this was that SST contains sentences that were more formal than Twitter. The differences between the embeddings' performances can provide insights into these embeddings. 

For this setting, we used the following embeddings: Word2Vec 400d trained on 400 million tweets, Word2Vec 300d trained on 6 billion tokens from Google News, GloVe 200d trained on 2 billion tweets and GloVe 200d trained on 6 billion words from Wikipedia and Gigaword5. Both of the Word2Vec embeddings were trained using skip-gram and negative sampling. We used the generated embeddings to conduct experiments on both SST and Twitter sentiment analysis. 

\subsection{Different Embeddings Trained on the Same Corpus}\label{dEsC}
Last but not least, we explored the different embeddings trained on the same corpus. Since different embedding algorithms would encode the information quite differently, JIVE analysis on those embeddings can provide us some insights on what are the shared and individual information. We used Word2Vec 400d and GloVe 200d both trained on Twitter to conduct experiment on Twitter sentiment analysis. 

\subsection{Neural Network Model}
We used the sipmle CNN model proposed by \citet{kim2014convolutional}. This model used CNN kernels of size $N$ by $L$, where $N$ was the kernel size and $L$ was the sentence length to simulate n-grams and sequential information of a sentence. In our case, we first padded all sentences to the max length and applied a kernel of size 5 to the sentences. 

\section{Empirical Results} \label{result}
In this section, we will present the results corresponding to section \ref{design}. First of all, all JIVE iterations are converged. 

\begin{figure}
    \centering
    \includegraphics[scale=0.5]{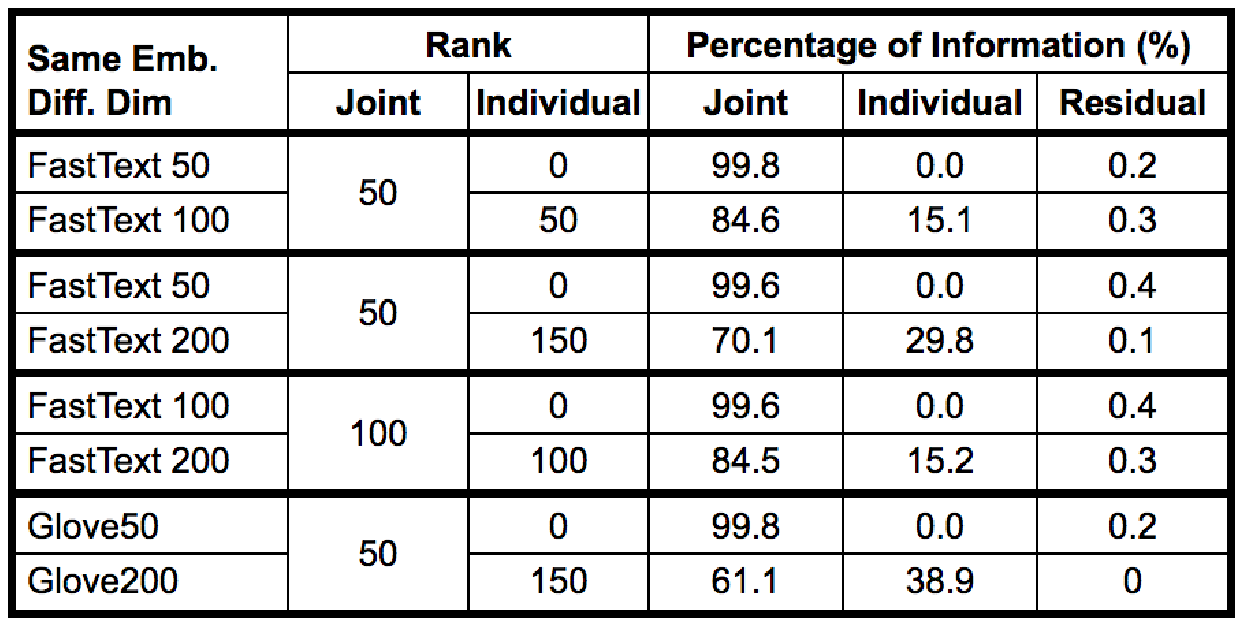}
    \caption*{Table 2. Variance explained percentage for the same embedding algorithm with different dimensions}
    \label{tab:sameembed_diffdim}
\end{figure}

\subsection{Setting 1: Same embedding different dimension - large dimension always captures }
In this section, we present the results of using same embedding method (FastText and GLoVE) with different number of dimension. First of all, among all settings discussed in section \ref{sEdD}, the embeddings with higher dimension can always capture all the information in lower dimensional embedding based on JIVE as shown in Table 2, which is as expected and also justifies that method works properly. However, the among of information stored in embeddings is not linearly increased with the number of dimension. For example, the 50 dimensional FastText takes 84.6\% information in 100 dimensional FastText. Table 5 summarizes the performance on SST-based sentiment analysis for different components. The joint component of rank 50 for GLoVE has a better performance than the original GLoVE with 50 features. Moreover, for FastText, the concatenated data between FastText50 and FastText200 achieved a higher performance than the original FastText200. These findings suggest that JIVE could help improve the performance under some situations.

\begin{figure}
    \centering
    \includegraphics[scale=0.5]{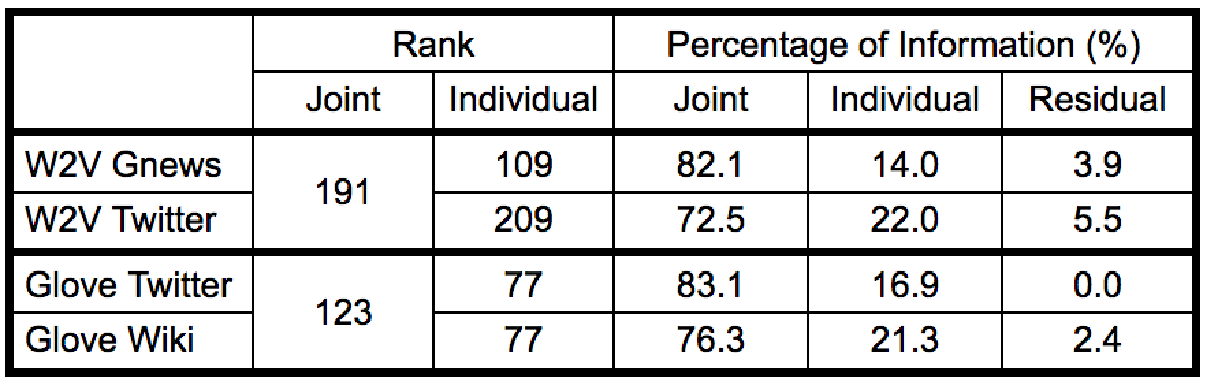}
    \caption*{Table 3. Variance explained percentage for the same embedding algorithm trained with different corpus}
\end{figure}

\begin{figure}
    \centering
    \includegraphics[scale=0.5]{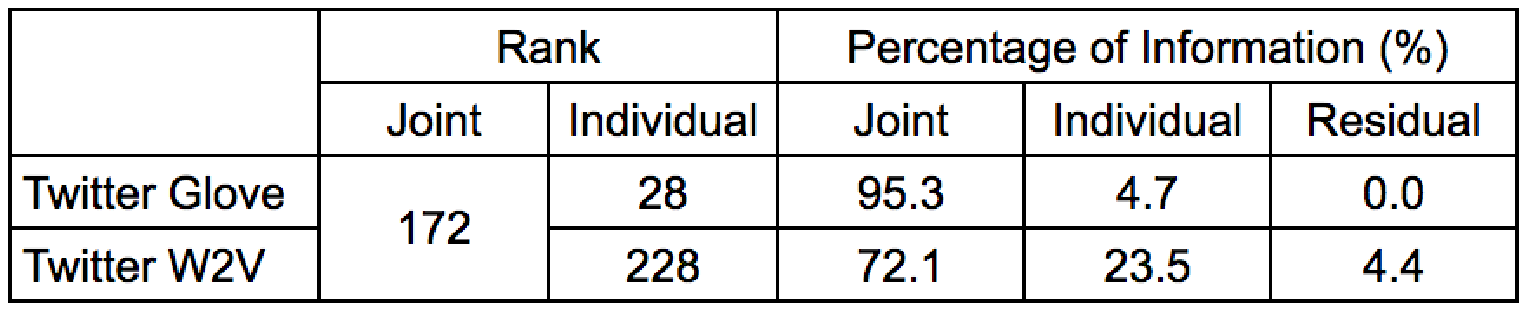}
    \caption*{Table 4. Variance explained percentage for the GLoVE and Word2Vec trained with on the same Twitter corpus}
\end{figure}

\begin{figure*}
    \centering
    \includegraphics[scale=0.6]{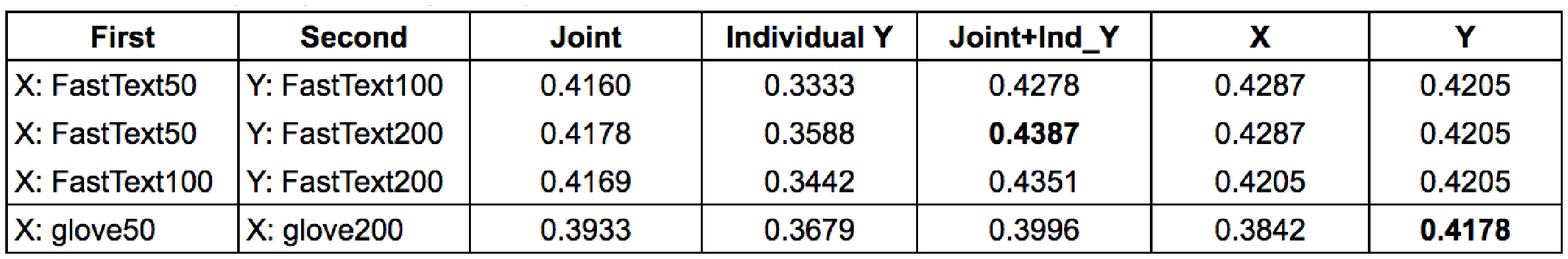}
    \caption*{Table 5. Accuracy on SST sentiment analysis for the same embedding and different dimensions}
    \label{tab:sameembed_diffdim}
\end{figure*}

\begin{figure*}
    \centering
    \includegraphics[scale=0.6]{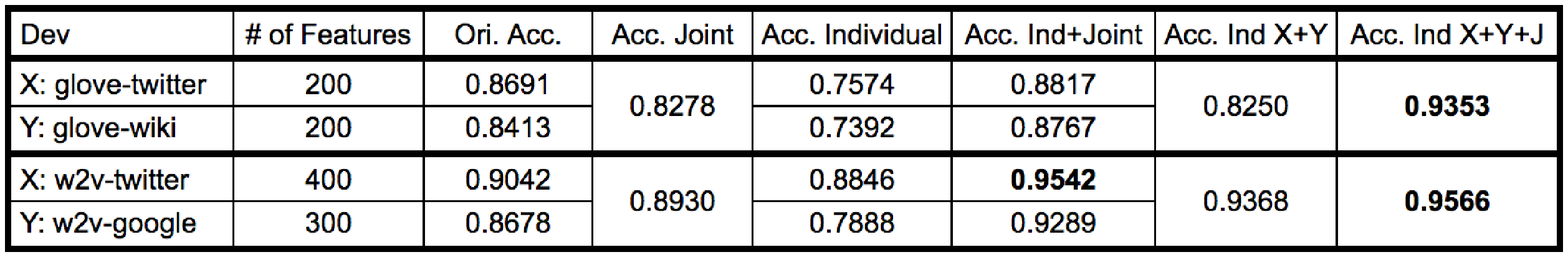}
    \caption*{Table 6. Accuracy on Twitter sentiment analysis for the GLoVE and Word2Vec trained on different corpus}
\end{figure*}

\subsection{Setting2: Same embedding trained on different corpus}

Table 3 summarizes the percentage of information captured by different components from word2vec and GLoVE trained on the different corpus. Although trained on different corpus, the information in joint component is higher than 70\% in all situations which indicates a strong similarity. But different from setting 1, all trained embedding methods still have a considerable amount of information in their individual components. For example, word2vec trained on Twitter corpus still keeps a 22.0\% information in the individual component. This indicates that different corpus did contribute to the final estimated embeddings significantly. 

For sentiment analysis, Table 6 summarizes the results on Twitter sentiment analysis for GLoVe trained on Twitter or Wiki corpus and for Word2Vec trained on Twitter and Google News corpus. First of all, for the original trained embedding (before JIVE), embeddings trained on Twitter corpus achieved a higher accuracy for Twitter sentiment analysis which is as expected. The joint component for word2vec trained on Twitter and Google News has a better accuracy than original performance of word2vec on Google News corpus. Although the joint part has slightly lower accuracy than original for word2vec on Twitter, the concatenated data (Joint+Ind) for word2vec-Twitter achieved a much higher accuracy than its original one (by 5\%). Similar for GLoVE on Twitter and Wiki, the (Joint+Ind) for GLoVE-Twitter has a better accuracy than its original performance. In either cases, the final concatenated (Ind X+Y+ Joint) can always provide much higher sentiment classification accuracy, which indicates the possibility of generating better word embedding by JIVE-based concatenation. We also note that Wiki corpus has relatively lower quality compared with Google News. This might be reason that the joint component for GLoVE-Wiki and GLove-Twitter has worse accuracy than their orginal ones. 

Moreover, we also included the results on SST sentiment analysis for word2vec trained on Twitter and Google News corpus in Table 7 which serves as a further sanity check since both embeddings are trained from the different corpus than the corpus used for testing. We find similar results where the joint component improves the original ones with lower accuracy.

\begin{figure*}
    \centering
    \includegraphics[scale=0.6]{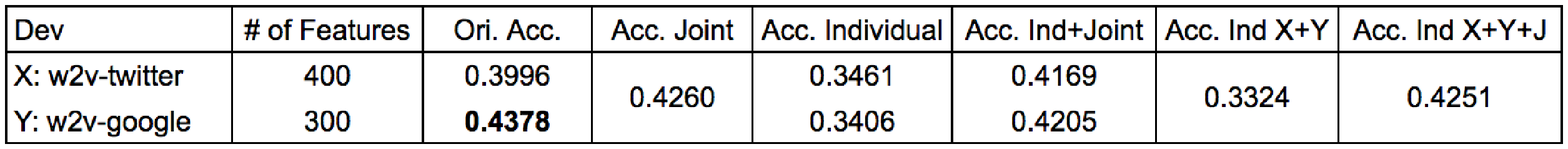}
    \caption*{Table 7. Accuracy on SST sentiment analysis for the Word2Vec trained on Twitter and Google News corpus}
\end{figure*}

\begin{figure*}
    \centering
    \includegraphics[scale=0.6]{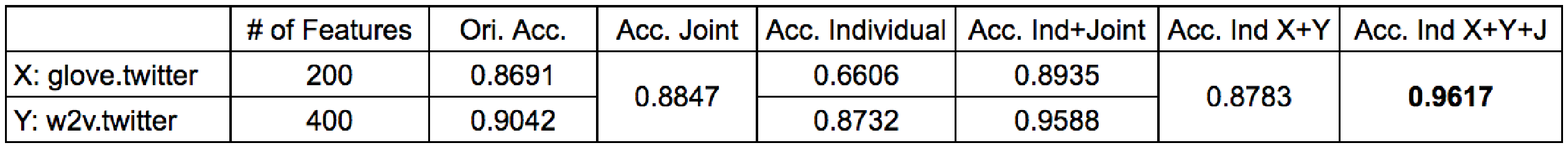}
    \caption*{Table 8. Accuracy on Twitter sentiment analysis for the GLoVE and Word2Vec trained on Twitter corpus}
    \label{tab:sameembed_diffdim}
\end{figure*}

\subsection{Setting3: Different embeddings trained on the same corpus}

Table 4 summarizes the percentage of information captured by different components from word2vec and GLoVE trained on the same Twitter corpus. The joint rank of GLoVE and Word2Vec trained on Twitter data is 172, and although trained by different algorithm, the joint component captures 95.3\% information for GLoVE. This finding indicates that corpus is very important in training word embedding method. Table 8 summarizes the performance on Twitter sentiment analysis for different components. Again, the joint component has higher accuracy than the original lower one - GLoVE-Twitter, and the JIVE-based concatenated embeddings always have better performance than their original ones, which further justifies the usage of JIVE in analyzing different word embeddings.

\section{Discussion and Conclusion} \label{conclu}

In this paper, we analyzed the trained word embeddings from fastText, word2vec and Glove with different dimensionalities and corpus based on JIVE. With the help of JIVE method, we decompose different word embedding results into joint and individual components. Based on sentiment analysis on 2 public available datasets, we evaluate the information contained in different part of the word embedding. We found that for the same embedding algorithm trained with the same corpus, higher dimensional embeddings can always capture all the information in the lower dimensional embeddings but the amount of information is not increased linearly with dimension. This finding justifies the mechanism of word embedding algorithm and provides hints for selection of dimensionality by checking the amount of information captured in different cases. Furthermore, we find that corpus and embedding algorithm all contribute to the final trained embeddings significantly where the individual component always capture considerable amount of information. However, in many situations, by mapping 2 word embeddings into the same space, the joint component can improve the original one with lower performance even with a smaller dimension in the Joint component. This finding is very interesting and provides a possibility for improve the performance of a word embedding. Finally, based on Twitter sentiment analysis, JIVE-based concatenated word embeddings always have a better performance than their original ones. This suggests 2 things: 1) JIVE can find important features and remove noises in word embedding; 2) combining different word embeddings could generate a better and more informative word embeddings. Overall, our work provides great insights for understanding the similarity and difference of different word embedding algorithms from a numerical perspective, and provides many opportunities for researchers in NLP field to further study and improve the word embeddings.

\section{Limitation and Future Work} \label{future}

Currently, we only evaluated the performance based on sentiment analysis. In the future, we plan to consider other types of NLP tasks such as sequence tagging (NER) to better evaluate the estimated joint and individual components from JIVE. Another limitation of the current work is on the embeddings. 

\bibliography{acl2019}
\bibliographystyle{acl_natbib}

\end{document}